\documentclass[letterpaper, 10 pt, conference]{ieeeconf}  

\IEEEoverridecommandlockouts                              

\usepackage{amsmath}
\usepackage{mathtools,amssymb}
\usepackage{hyperref}
\usepackage{bm}
\usepackage{color}
\usepackage{xcolor}
\usepackage{cite}
\usepackage{booktabs}
\usepackage{multirow}
\usepackage[linesnumbered,ruled,vlined]{algorithm2e}
\usepackage[binary-units]{siunitx}
\usepackage{subcaption}
\usepackage{makecell}
\usepackage{graphicx}
\usepackage{balance}

\usepackage{mwe,tikz}

\newcommand{\algrule}[1][.4pt]{\par\vskip.2\baselineskip\hrule height #1\par\vskip.2\baselineskip}

\newcommand{\Rthree}[0]{\mathbb{R}^3}
\newcommand{\SOthree}[0]{\mathbb{SO}(3)}

\newcommand{\disable}[1]{}

\DeclareMathOperator*{\minimize}{\text{minimize}}
\DeclareMathOperator*{\subjectto}{\text{s.t.}}

\title{\LARGE \bf Iterative Active-Inactive Obstacle Classification for Time-Optimal Collision Avoidance}

\author{Mehmetcan Kaymaz$^{1}$, Nazim Kemal Ure$^{2}$
\thanks{The authors are with the Data \& Decision Lab.,  ITU Artificial Intelligence and Data Science Application
and Research Center, Istanbul Technical University, Turkey.}
\thanks{{\tt \small \{kaymazm16$^{1}$, ure$^{2}$\}@itu.edu.tr}}%
}

\renewcommand\subsubsection[1]{\vspace{0pt}\noindent\textbf{#1.}}

\begin{document}

\setlength{\abovedisplayskip}{6pt}
\setlength{\belowdisplayskip}{6pt}
\setlength{\abovedisplayshortskip}{4pt}
\setlength{\belowdisplayshortskip}{4pt}

\maketitle

\begin{abstract}
Time-optimal obstacle avoidance is a prevalent problem encountered in various fields, including robotics and autonomous vehicles, where the task involves determining a path for a moving vehicle to reach its goal while navigating around obstacles within its environment. This problem becomes increasingly challenging as the number of obstacles in the environment rises. We propose an iterative active-inactive obstacle approach, which involves identifying a subset of the obstacles as "active", that considers solely the effect of the "active" obstacles on the path of the moving vehicle. The remaining obstacles are considered "inactive" and are not considered in the path planning process. The obstacles are classified as 'active' on the basis of previous findings derived from prior iterations. This approach allows for a more efficient calculation of the optimal path by reducing the number of obstacles that need to be considered. The effectiveness of the proposed method is demonstrated with two different dynamic models using the various number of obstacles. The results show that the proposed method is able to find the optimal path in a timely manner, while also being able to handle a large number of obstacles in the environment and the constraints on the motion of the object.
\end{abstract}

\section{Introduction}

Collision avoidance is a critical topic in robotics and autonomous systems because of its direct impact on system's safety and efficiency. While there are several strategies for avoiding collisions in cluttered environments, time-optimal approaches are especially attractive since they allow the system to arrive at its target location in the shortest amount of time, while still avoiding collisions. 

Polynomial methods \cite{Richter2016}, sampling-based methods \cite{Webb2013} \cite{penicka22RALmintimeplanning}, optimization-based methods \cite{foehn2021CPC}, searching-based methods \cite{Liu2017}, and learning-based methods \cite{penicka22learning} are some of the predominant approaches for addressing time-optimal collision avoidance in robotics. Recent studies by Penicka et al. \cite{penicka22RALmintimeplanning} and \cite{penicka22learning} have illustrated that optimization-based approaches consistently outperform alternative methodologies in terms of solution quality. However, it is worth noting that the computational demands associated with optimization-based techniques can become prohibitively high, particularly as the number of obstacles in the environment increases.

This computational challenge arises from the fact that each obstacle introduces non-convex constraints into the optimization problem, significantly augmenting its overall complexity \cite{foehn2021CPC}. Consequently, existing solvers occasionally encounter difficulties in converging to a valid solution when confronted with scenarios involving a large number of obstacles.

This study presents a pioneering methodology aimed at achieving time-optimal collision avoidance within intricate, cluttered environments. The approach hinges upon the categorization of obstacles into active and inactive entities, thereby permitting the resolution of optimization challenges exclusively for the former. Drawing upon prior findings, the status of active obstacles undergoes iterative refinement. Our method was subjected to rigorous assessment across multiple dynamic models featuring diverse obstacle configurations. The results unequivocally demonstrate the superior performance of the proposed approach compared to conventional method. In the evaluation of three key performance metrics, namely computation time, trajectory duration, and success rate, our algorithm demonstrated superior performance compared to traditional approaches, particularly in terms of success rate. In our simulation involving a quadrotor, the mean success rate achieved by the classical approach was 40\%, whereas our proposed method yielded a significantly enhanced mean success rate of 73\%.

\section{Related Work}\label{sec:related}

Contemporary methodologies for achieving agile quadrotor flight predominantly hinge upon conventional strategies that compartmentalize the processes of trajectory planning and control, with a predominant emphasis on enhancing trajectory smoothness as opposed to the pursuit of optimal minimum-time flight. In the context of evolving technological landscapes, there has been a discernible surge in interest surrounding learning-based techniques, chiefly attributable to the burgeoning prominence of neural networks within the domain. Among these emerging paradigms, several have garnered noteworthy attention and recognition.

Polynomial methods \cite{Richter2016, burri2015real-time, han2021fastracing, mueller2015TRO_minjerk} and B-spline representations \cite{Zhou2020, zhou2020raptor, Penin2018RAL_vision_reactive_planning} are frequently employed in the field of quadrotor trajectory planning due to their exploitation of the differential flatness property inherent in quadrotor dynamics, which enables the generation of trajectories with a strong emphasis on smoothness. However, a notable limitation of these approaches arises from their inherent bias toward producing smooth trajectories, which renders them suboptimal for scenarios where the primary objective is to minimize flight time. This limitation stems from their inability to effectively represent rapid and abrupt changes in the quadrotor's state or input commands, which are often essential for achieving minimum-time flight profiles.

Search-based methods, as described in \cite{Liu_search_based_LQMTC} and \cite{liu2018search}, transform trajectory planning into a computationally intensive graph search problem where the primary objective is to optimize the trajectory with respect to time while adhering to predefined discretization bounds. Nevertheless, these techniques encounter a significant challenge known as the "curse of dimensionality," which stems from the exponential growth in computational complexity as the dimensionality of the search space increases. Furthermore, they often rely on simplifications such as employing point-mass models, which, although efficient, impose limitations on their applicability, particularly when dealing with the intricacies of full quadrotor dynamics.

Sampling-based kinodynamic methods, as extensively discussed in \cite{Webb2013} and \cite{penicka22RALmintimeplanning}, operate by incrementally expanding state trees through the utilization of randomized state and input sampling strategies. However, a notable limitation inherent to these techniques is their inherent propensity to prioritize the minimization of trajectory length rather than the more critical objective of minimizing time. This predilection for trajectory length optimization can be particularly disadvantageous in scenarios involving quadrotor dynamics, where the intricate and challenging nature of the system dynamics renders traditional sampling-based planners less suitable for achieving minimum-time objectives.

Optimization-based approaches have gained prominence in addressing motion planning challenges by formulating the problem as constrained optimization tasks \cite{foehn2021CPC, li2023, small2019NPMCobstalces, Garimella17NMPCobstacles, Lindqvist20NMPCobstacles}. These methodologies offer the distinct advantage of accommodating nonlinear dynamical systems and enforcing intricate constraints within the planning framework. Nevertheless, it is imperative to acknowledge that such methods often exhibit extended computational times, primarily due to the complexity of solving optimization problems, and they occasionally resort to approximations to alleviate the computational burden. Additionally, one notable limitation is their potential inefficiency in handling obstacles within the environment, as optimization processes may require considerable iterations to navigate around or avoid obstacles effectively.

Emergent challenges in the field of aerial robotics have spurred the development of innovative learning-based methodologies \cite{penicka22learning, Loquercio21FlightWild, kaufmann2020RSS, Fuchs21RLgrandTurismo, song21RLdroneRacing, kaufmann2022benchmark}. These methodologies, predominantly characterized by the training of neural network policies tasked with predicting control commands from high-dimensional sensor observations, offer promising avenues for achieving agile flight in complex environments. However, it is essential to acknowledge that while these approaches exhibit remarkable capabilities, they often demand substantial data collection efforts and may grapple with scalability constraints.

In the broader context of the scholarly discourse, a persistent and unresolved challenge pertains to the efficacious resolution of the time-optimal collusion avoidance problem, particularly when confronted with a substantial proliferation of obstacles. Notably, optimization-based algorithms have demonstrated exceptional utility in ascertaining optimal solutions to this predicament; however, their efficacy falters precipitously in the face of an escalating abundance of obstacles. In response to this exigent issue, our proposed algorithm proffers a substantial enhancement to the performance of optimization-based methodologies, thereby ameliorating the time-optimal collision avoidance conundrum.

\subsection{Quadcopter Dynamics}

The states of quadrotor are $\bm{x_{quad}}=\begin{bmatrix} \bm{p},\bm{q},\bm{v},\bm{\omega} \end{bmatrix}^{T}$ consists of its position $\bm{p} \in \Rthree$, velocity $\bm{v} \in \Rthree$, unit quaternion rotation $\bm{q} \in \SOthree$, and body rates $\bm{\omega} \in \Rthree$. 
The total collective thrust $\bm{f}_{T}$ and body torque $\bm{\tau}$ are inputs. The dynamic equations are
\begin{align}
\label{eq:quat_dyn}
  \begin{aligned}
    \bm{\dot{p}} &= \bm{v} \vphantom{\frac{1}{2}} \\
    \bm{\dot{v}} &= \frac{1}{m}R(\bm{q})\bm{f}_{T} + \bm{g}
  \end{aligned}
  &&
  \begin{aligned}
    \bm{\dot{q}} &= \frac{1}{2} \bm{q} \odot \begin{bmatrix} 0 \\ \bm{\omega} \end{bmatrix} \\
    \bm{\dot{\omega}} &= \bm{J}^{-1} (\bm{\tau} - \bm{\omega} \times \bm{J} \bm{\omega}) \vphantom{\frac{1}{2}} \text{,}
  \end{aligned}
\end{align}
where $\odot$ denotes quaternion multiplication, $R(\bm{q})$ quaternion rotation, $m$ quadcopter mass, $\bm{J}$ its inertia, and $\bm{g}$ is gravity.

However, the real quadrotor inputs are single rotor thrusts $\begin{bmatrix}f_1,f_2,f_3,f_4\end{bmatrix}$ which are used to calculate $\bm{f}_{T}$ and $\bm{\tau}$ as
\begin{equation}
\label{eq:tau_thrust}
\bm{f}_{T}=\begin{bmatrix} 0 \\ 0 \\ \sum f_i \end{bmatrix} \text{, }
\bm{\tau}_{b} = 
\begin{bmatrix} 
 l/\sqrt{2}(f_{1}-f_{2}-f_{3}+f_{4})   \\
 l/\sqrt{2}(-f_{1}-f_{2}+f_{3}+f_{4})   \\
 \kappa (f_{1}-f_{2}+f_{3}-f_{4})
\end{bmatrix} 
\text{,}
\end{equation}
using torque constant $\kappa$ and arm length $l$.
The single rotor thrusts are further constrained~\eqref{eq:motor_constraints} by minimal $f_{min}$ and maximal $f_{max}$ values.
The body rates are limited~\eqref{eq:rate_constraints} by a per-axis maximal allowed value $\omega_{max}$.
\begin{align}
  f_{min} \leq &f_{i} \leq f_{max} \text{, for } i \in \{1,\ldots,4\} \label{eq:motor_constraints}\\
  -\omega_{max} \leq &\omega_{i} \leq \omega_{max} \text{, for } \bm{\omega}={\begin{bmatrix}\omega_{1},\omega_{2},\omega_{3}\end{bmatrix}}^{T} \label{eq:rate_constraints}
\end{align}

\subsection{Time-Optimal Collusion Avoidance}
Let's say $o_i$ is a single obstacle and $O=\begin{bmatrix} o_1, o_2,...,o_{N_{obs}} \end{bmatrix}$ is the space covered by all obstacles. Assume $C_{free}$ is the collision-free area. Let's say $x_{initial}$ is the initial state and $x_{final}$ is the final state. The time-optimal collision avoidance problem is defined as

\begin{equation} \label{objective}
   \begin{split}
      \minimize_{X,U,T_f}~& T_f    \\
      \subjectto \text{ }& \bm{p_i} \in C_{free} \\
      &  \bm{x_{i+1}}  = f_{dyn}(\bm{x_i},\bm{u_i},T_f/N) \text{ for } i \in \{0,\ldots,N\}\text{,}\\
      & \bm{x_0}= x_{initial}, \bm{x_{N}}=x_{final} \\
      & T_f/N <= dt_{max} \\
      &  \eqref{eq:motor_constraints}\text{, } \eqref{eq:rate_constraints} \text{.}
   \end{split}
\end{equation}

where $f_{dyn}(.)$ is the discreate-time dynamic model obtained with RK4, \autoref{eq:quat_dyn} and \autoref{eq:tau_thrust}, $N$ is the lenght of the trajectory and $dt_{max}$ is the maximum step size for dynamic model. The optimization problem defined in \autoref{objective} aims to find a feasible trajectory while minimizing time. The problem is non-convex because of the collusion avoidance constraint and quadrotor dynamics. Let's assume $o$ is a circle positioned in $p_o$ with radius $r_o$. The $\bm{p_i} \in C_{free}$ constraint can be written as 

\begin{align} \label{constraint-circle}
    ||p_i-p_o||>r_o + \epsilon
\end{align}

for one obstacle. $\epsilon$ is the sum of the quadrotor and safety distance. The \autoref{constraint-circle} must be written for all obstacles individually. Also, \autoref{constraint-circle} is a non-convex constraint so, the non-convexity of the problem increases with the number of obstacles. Therefore, solvers fail to solve \autoref{objective} when the number of obstacles is high.

\section{Methodology\label{sec:method}}

\begin{figure*}[!ht]
    \centering
    \includegraphics[width=.8\textwidth]{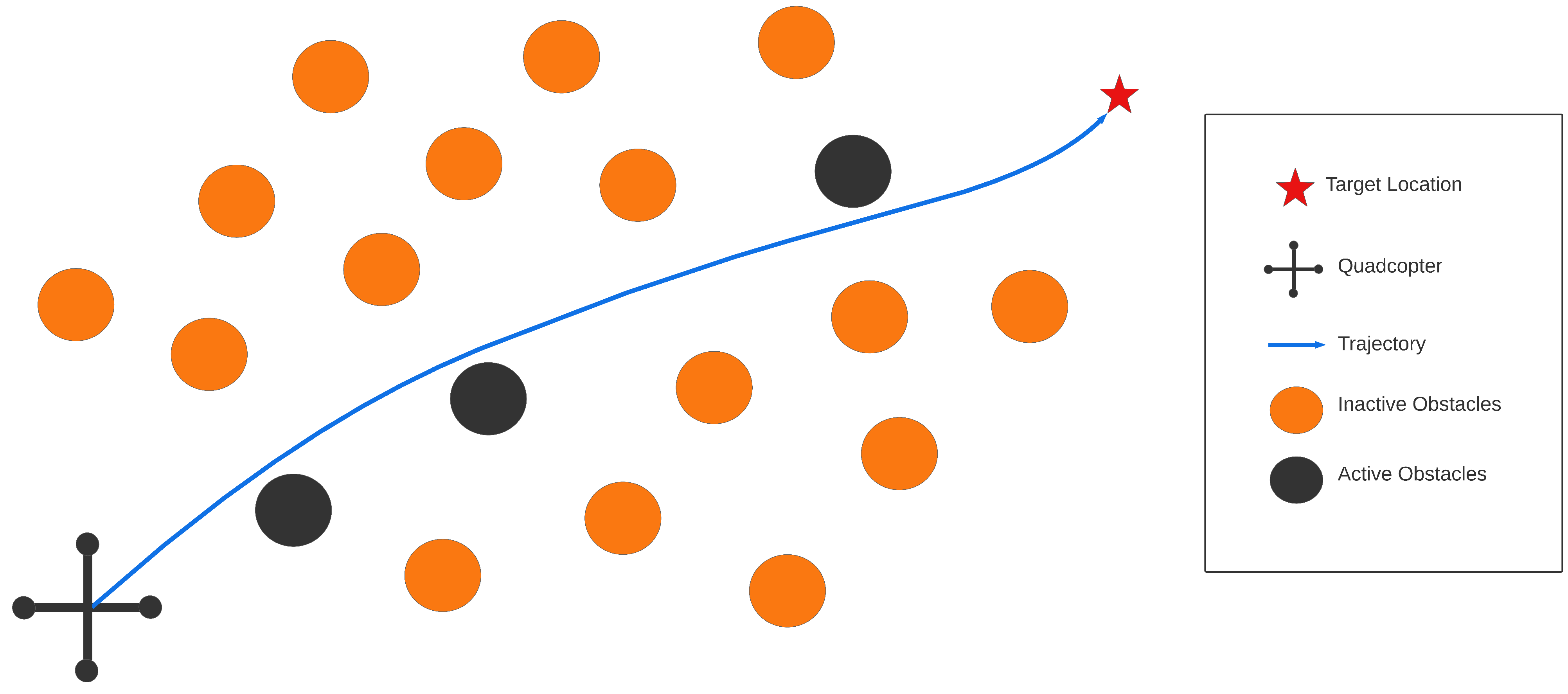}
    \caption{Visualization of the proposed method. The obstacles in the scenario are categorized into two distinct sets: the 'active' obstacles, depicted in black, and the 'inactive' obstacles, represented in orange. The optimization problem, as defined in \autoref{objective}, is exclusively addressed with respect to the 'active' obstacles. Consequently, the resulting solution, denoted by the blue line, is derived solely from the constraints imposed by the 'active' obstacles.}
    \label{fig:mainfig}
\end{figure*}

The proposed approach is founded on the premise that not all obstacles must be taken into account when addressing the optimization problem articulated in \autoref{objective}. By reducing the consideration of obstacles, we mitigate the non-convex nature of the optimization problem, rendering it more amenable to solution by optimization solvers. This method's conceptual framework is visually represented in \autoref{fig:mainfig}. The solution of a time-optimal trajectory closely involves only three specific obstacles, eliminating the need to consider other obstacles. It is, however, imperative to underscore that there is currently no mechanism in place for discerning which obstacles warrant consideration and which do not.

To address this issue, we have devised a strategy wherein obstacles are categorized as either "active" or "inactive." Concretely, let $O_{active}=[o_1,o_2,...,o_{N_{active}}]$ denote the set of active obstacles, and $O_{inactive}=[o_1,o_2,...,o_{N_{inactive}}]$ denote the set of inactive obstacles. This categorization adheres to the constraint $O=O_{active} \cup O_{inactive}$, with $N_{obs}=N_{active}+N_{inactive}$. Consequently, we can suppose that the solution $X=\begin{bmatrix} \bm{x_0},\bm{x_1},...\bm{x_N} \end{bmatrix}$, which satisfies the optimization problem delineated in \autoref{objective}, pertains exclusively to the active obstacles ($O_{active}$).

In light of this categorization, we have established a feasibility check algorithm, the specifics of which are outlined as follows:

\SetAlFnt{\footnotesize}
\setlength{\textfloatsep}{0pt}
\begin{algorithm}[!htb]
   \caption{Feasibility Check\label{alg:feasibility}} 
   \SetKwInOut{Input}{Input}
   \SetKwInOut{Output}{Out}
   \Input{$X$, $O_{inactive}$} 
   \Output{is\_feasible, $O_{new}$}
   \algrule
   is\_feasible $\leftarrow$ true \\
   $O_{new} \leftarrow \emptyset$ \\
   \DontPrintSemicolon
   \ForEach{$ \bm{x_i} \in X $}{
    \ForEach{$o_j \in O_{inactive}$}{
    \If{$||\bm{p_i}-p_{o_j}||<=r_{o_j} + \epsilon$ and $o_j \notin O_{new}$}{ $O_{new}$.add($o_j$) \\
    is\_feasible $\leftarrow$ false
    }
    }
   }
\vspace{-0.5em}
\end{algorithm}

The algorithm denoted as \autoref{alg:feasibility} serves the purpose of verifying the feasibility of trajectory $X$ and subsequently identifying the specific obstacles that compromise the trajectory's feasibility. This iterative active-inactive obstacle methodology, employed in tandem with the feasibility assessment, encompasses the broader framework.

\SetAlFnt{\footnotesize}
\setlength{\textfloatsep}{0pt}
\begin{algorithm}[!htb]
   \caption{Iterative Active-Inactive Obstacle Approach \label{alg:method}} 
   \SetKwInOut{Input}{Input}
   \SetKwInOut{Output}{Out}
   \Input{$O$, $x_{initial}$, $x_{final}$, $N$, $dt_{max}$} 
   \Output{solved, $X$, $T_f$}
   \algrule
   solved $\leftarrow$ false \\
   $O_{inactive} \leftarrow O$ \\
   $O_{active} \leftarrow \emptyset$ \\
   \DontPrintSemicolon
    \While{!solved}{
    $X$, $T_f$=Solve($O_{active},x_{initial},x_{final}$,$N$,$dt_{max}$) Eq. \ref{objective} \\
    is\_feasible, $O_{new}$=FeasibilityCheck($X$,$O_{inactives}$) Alg.\ref{alg:feasibility} \\
    \If{is\_feasible}{
    solved $\leftarrow$ true \\
    }
    \Else{
    $O_{actives}$.add($O_{new}$) \\
    $O_{inactives}$.remove($O_{new}$) \\
    }
    }
\vspace{-0.5em}
\end{algorithm}

The algorithm initially sets all obstacles to an inactive state, signifying the absence of any active obstacles during the first iteration. Subsequently, it computes the trajectory denoted as $X$ by solving the optimization problem described in \autoref{objective}. The algorithm then assesses the feasibility of this computed trajectory. If the trajectory is found to be infeasible, the obstacles responsible for compromising its feasibility are identified and transitioned from the inactive obstacle set to the active obstacle set.

In the event that the trajectory $X$ is determined to be feasible, it is considered as the solution to the problem. It is important to note that there is no collision between the trajectory $X$ and the active obstacles, a property ensured by the optimization objective defined in \autoref{objective}. Furthermore, there is also no collision between the trajectory $X$ and the inactive obstacles, as guaranteed by the constraints specified in \autoref{alg:feasibility}. Consequently, the output of the algorithm described in \autoref{alg:method} in the form of trajectory $X$ represents a collision-free path in the presence of all obstacles.


Proving the efficacy of employing \autoref{alg:method} to enhance solution outcomes presents a formidable challenge due to the inherent non-convexity of the underlying optimization problem described in \autoref{objective}. In non-convex problems, it is notoriously difficult to establish proofs of global optimality.

Consider the establishment of two distinct feasible sets: one denoted as $D$, encompassing the inclusion of all obstacles $O$ in the optimization process defined within \autoref{objective}, and another, referred to as $D'$, that only takes into account the active obstacles subset, denoted as $O_{active}$, within \autoref{objective}. It is worth noting that $O_{active}$ is a proper subset of the complete set of obstacles, represented as $O$ ($O_{active} \subset O$), thus leading to the natural inclusion of $D$ within $D'$ ($D \subset D'$), as the incorporation of more obstacles within $O$ naturally constrains the feasible solution space.

Now, let's assume, hypothetically, that \autoref{objective} conforms to the realm of convex optimization problems. In such a case, if we find a solution, represented by $X', U', T_f'$, while operating within the scope of the feasible set $D'$, we can unequivocally assert that this solution represents the global minimum, owing to the convexity of the problem. Importantly, if $X', U', T_f'$ also lie within the bounds of the more extensive feasible set $D$, we can confidently extend this conclusion to assert that it also constitutes the optimal solution when considering the entire obstacle set, as $D$ is inherently contained within $D'$.

However, it is essential to acknowledge that the problem formulated in \autoref{objective} is not, in fact, a convex optimization problem. This crucial distinction introduces an element of uncertainty into our analysis. In particular, when dealing with non-convex optimization, the solution $X', U', T_f'$ obtained within the confines of $D'$ may not necessarily correspond to the global minimum. This inherent non-convexity introduces the possibility that solving \autoref{objective} for $D'$ might lead to a local minimum, while a broader exploration of the problem space encompassing $D$ could reveal a superior global minimum.

Nevertheless, it is worth noting that, in our extensive simulations and empirical observations, we have not encountered any scenarios in which the solution derived from the optimization problem considering the entire obstacle set, denoted as $D$, surpasses the solution derived from the limited scope of $D'$. These findings, while not constituting formal proof, do provide empirical evidence suggesting that, in practice, the inclusion of all obstacles in the optimization process, as opposed to just the active obstacles, generally leads to more favorable outcomes.

\section{Results\label{sec:results}}

The proposed algorithm is tested on the forest scenario with a single target problem. The trajectory is between the start and target states. To see the general performance of the proposed method, we also added a point-mass dynamic system in the 2D environment to our experiments. The algorithm tested on both a 2D point-mass model and a 3D quadrotor model with varying numbers of obstacles. In the 2D scenario, the obstacle is defined as a circle with a random radius between 0.1 and 0.2 meters in a random position in the environment. In the 3D scenario, the obstacle is defined as a cylinder with a random radius between 0.1 and 0.2 meters in a random position. The height of the cylinder is equal to the height of the environment so the quadrotor can not go over the obstacles. The collision avoidance constraint for both scenarios is the same without the height effect, the cylinder is the same as the circle for the collision avoidance constraint. The constraint for both scenarios is given in \autoref{constraint-circle} but in 3D scenarios, the collusion avoidance constraint is written for only $x-y$ positions. For the 2D scenario, the environment size is 10-10 meters. The initial state is $x_{initial}=\begin{bmatrix} 0,0,0,0 \end{bmatrix}^T$ and the final state is $x_{final}=\begin{bmatrix} 10,10,0,0 \end{bmatrix}^T$. For the 3D scenario, the environment size is 10-10-10 meters. The initial state is $x_{initial}=\begin{bmatrix} 0,0,5,1,0,0,0,0,0,0,0,0,0,0 \end{bmatrix}^T$ and the final state is $x_{final}=\begin{bmatrix} 10,10,5,1,0,0,0,0,0,0,0,0,0,0 \end{bmatrix}^T$. The parameters in the simulation are given in \autoref{tab:config}.

\vspace{0.5em}
\begin{table}[!htb] 
   \centering 
   \footnotesize 
   \renewcommand{\tabcolsep}{1.4pt} 
   \renewcommand{\arraystretch}{0.8} 
   \caption{Quadrotor parameters\label{tab:config}} 
   \begin{tabular}{cc}
     \toprule 
   Variable & Value \\
       \midrule
     $m$ [\SI{}{\kg}] & 0.85   \\
     $l$ [\SI{}{\meter}] & 0.15 \\
    $f_{min}$ [\SI{}{\newton}] & 0   \\
    $f_{max}$ [\SI{}{\newton}] & 7 \\
    $\text{diag}(J)$ [\SI{}{\gram\metre\squared}] & $[1,1,1.7]$  \\
     $\kappa$  & 0.05 \\
    $w_{max}$ [\SI{}{\radian\per\second}] & 15  \\[-0.2em] 
\bottomrule 
   \end{tabular}  
   \vspace{1.0em}
\end{table}

We modified CPC \cite{foehn2021CPC} by adding a position constraint and compared it with our proposed method which is equivalent to solving \autoref{objective} with only one waypoint. There are other methods for time-optimal collision avoidance but \cite{penicka22RALmintimeplanning} and \cite{penicka22learning} showed that if CPC finds a solution, it is superior to others. We validate both CPC and our algorithm in environments that include 5, 10, 15, 20, 25, 30, 40, 50, 60, 70, 80, 90, and 100 obstacles for both scenarios. For each number of obstacles, we simulated 100 random scenarios and reported mean performance parameters. The IPOPT\cite{ipopt} solver is used to solve \autoref{objective}. The results for the 2D scenario are seen \autoref{tab:pointmassres}. Our proposed method outperformed CPC in terms of computation requirement, final time of the trajectory, and success rate. The computational requirement of our algorithm is less than CPC because the non-convex collision avoidance constraint increases the complexity of the problem which causes lots of computation requirements. For example, imagine a scenario that includes 30 obstacles. the CPC directly solves the problem for 30 obstacles. On the other hand, our algorithm tries 0 obstacles first and goes on until it finds a collusion-free trajectory. In general, the algorithm finds collusion-free trajectories with a much smaller number of obstacles like 4-5 in 2-3 iterations. Therefore, the total computation requirement for 2-3 iterations is much less than a single solution for all obstacles. The final time of the trajectory is the cost of the optimization problem defined in \autoref{objective}. As mentioned, each obstacle increases the non-convexity of the problem with causes more local minima. Therefore, solving the problem with less number of obstacles increases the probability of finding global minima so, the solution obtained from our algorithm has a better chance of being global minima as seen in the results. Similarly, the increase in non-convexity causes solvers not to converge and fail. That is why, our algorithm is also better because with fewer obstacles, the chance for the solver to fail decreases.

\begin{table*}
    \centering
    \footnotesize 
    {%
    \renewcommand{\tabcolsep}{5.6pt} 
    \renewcommand{\arraystretch}{0.8} 
    \caption{The outcomes pertaining to point-mass simulations across various obstacle quantities. Each scenario involved the assessment of 100 randomly generated obstacles. }\label{tab:pointmassres}}
    \begin{tabular}{ ccccccc }
        \toprule
        \multirow{2}{4em}{$n_{obs}$} & \multicolumn{3}{|c|}{CPC \cite{foehn2021CPC}}  & \multicolumn{3}{|c|}{\textbf{Ours}} \\
        
        & c. time[s] & $t_f$[s] & success[\%] & c. time[s] & $t_f$[s] & success[\%]\\
        \midrule
        5 & 14.19 $\pm$ 2.65  & 2.01 $\pm$ 0.01 & \textbf{100}  & \textbf{5.25 $\pm$ 0.74}  & \textbf{2.00 $\pm$ 0.00} & \textbf{100} \\
        \midrule
        10 & 31.44 $\pm$ 4.51  & 2.02 $\pm$ 0.03 & 99  & \textbf{8.10 $\pm$ 5.79}  & \textbf{2.00 $\pm$ 0.00} & \textbf{100} \\ 
        \midrule
        15 & 59.29 $\pm$ 30.68  & 2.04 $\pm$ 0.04 & 97  & \textbf{10.51 $\pm$ 2.56}  & \textbf{2.01 $\pm$ 0.00} & \textbf{100} \\
        \midrule
        20 & 95.88 $\pm$ 7.32  & 2.04 $\pm$ 0.04 & 90  & \textbf{15.30 $\pm$ 3.69}  & \textbf{2.01 $\pm$ 0.01} & \textbf{100} \\
        \midrule
        25 & 141.73 $\pm$ 7.46  & 2.06 $\pm$ 0.07 & 82  & \textbf{18.13 $\pm$ 3.92}  & \textbf{2.01 $\pm$ 0.01} & \textbf{100} \\
        \midrule
        30 & 184.17 $\pm$ 8.08  & 2.06 $\pm$ 0.06 & 67  & \textbf{19.05 $\pm$ 4.13}  & \textbf{2.01 $\pm$ 0.01} & \textbf{97} \\
        \midrule
        40 & 221.55 $\pm$ 8.31  & 2.07 $\pm$ 0.06 & 61  & \textbf{22.71 $\pm$ 4.40}  & \textbf{2.02 $\pm$ 0.02} & \textbf{99} \\
        \midrule
        50 & 320.11 $\pm$ 10.47  & 2.06 $\pm$ 0.06 & 47  & \textbf{26.52 $\pm$ 5.21}  & \textbf{2.02 $\pm$ 0.02} & \textbf{97} \\
        \midrule
        60 & 308.51 $\pm$ 10.37  & 2.10 $\pm$ 0.08 & 27  & \textbf{52.35 $\pm$ 7.29}  & \textbf{2.03 $\pm$ 0.03} & \textbf{97} \\ 
        \midrule
        70 & 409.69 $\pm$ 10.56  & 2.07 $\pm$ 0.05 & 13  & \textbf{51.25 $\pm$ 5.46}  & \textbf{2.03 $\pm$ 0.03} & \textbf{92} \\
        \midrule
        80 & 429.92 $\pm$ 11.80  & 2.05 $\pm$ 0.03 & 15  & \textbf{61.34 $\pm$ 6.20}  & \textbf{2.04 $\pm$ 0.03} & \textbf{89} \\
        \midrule
        90 & 512.38 $\pm$ 11.14  & 2.06 $\pm$ 0.06 & 10  & \textbf{84.45 $\pm$ 8.55}  & \textbf{2.04 $\pm$ 0.04} & \textbf{75} \\
        \midrule
        100 & 614.42 $\pm$ 12.12  & 2.06 $\pm$ 0.07 & 2  & \textbf{92.34 $\pm$ 9.55}  & \textbf{2.04 $\pm$ 0.05} & \textbf{69} \\
        \bottomrule
    \end{tabular}
\end{table*}

The results for 3D scenarios with the quadrotor model are seen in \autoref{tab:quadres}. Similar to 2D scenario results, our algorithm outperforms CPC in all metrics. The quadrotor is a much more complex system with respect to the point-mass model. Therefore, the CPC algorithm fails for all 100 random scenarios after 50 obstacles. On the other hand, Our proposed method solves 20 of 100 random scenarios even for 100 obstacles. The results also show us the effect of non-convexity level on solutions. As both results show, the solver starts to fail when the number of obstacles increases. 

\begin{table*}
    \centering
    \footnotesize 
    {%
    \renewcommand{\tabcolsep}{5.6pt} 
    \renewcommand{\arraystretch}{0.8} 
    \caption{The outcomes pertaining to quadrotor simulations across various obstacle quantities. Each scenario involved the assessment of 100 randomly generated obstacles.}
    \label{tab:quadres}}
    \begin{tabular}{ ccccccc }
        \toprule
        \multirow{2}{4em}{$n_{obs}$} & \multicolumn{3}{|c|}{CPC \cite{foehn2021CPC}}  & \multicolumn{3}{|c|}{\textbf{Ours}} \\
        
        & c. time[s] & $t_f$[s] & success[\%] & c. time[s] & $t_f$[s] & success[\%]\\
        \midrule
        5 & 59.17 $\pm$ 3.84 & 1.99 $\pm$ 0.01 & \textbf{100} & \textbf{48.98 $\pm$ 3.98} & \textbf{1.98 $\pm$ 0.00} & \textbf{100} \\
        \midrule
        10 & 123.12 $\pm$ 6.76 & 1.99 $\pm$ 0.01 & 97 & \textbf{56.89 $\pm$ 4.56} & \textbf{1.98 $\pm$ 0.00} & \textbf{100} \\
        \midrule
        15 & 150.79 $\pm$ 6.84 & 1.99 $\pm$ 0.01 & 92 & \textbf{61.76 $\pm$ 4.82} & \textbf{1.98 $\pm$ 0.00} & \textbf{98} \\
        \midrule
        20 & 176.92 $\pm$ 7.11 & 1.99 $\pm$ 0.01 & 75 & \textbf{82.17 $\pm$ 6.04} & \textbf{1.98 $\pm$ 0.00} & \textbf{93} \\
        \midrule
        25 & 264.29 $\pm$ 12.22 & 1.99 $\pm$ 0.01 & 62 & \textbf{91.20 $\pm$ 7.42} & \textbf{1.98 $\pm$ 0.00} & \textbf{89} \\
        \midrule
        30 & 369.00 $\pm$ 16.56 & 2.00 $\pm$ 0.03 & 48 & \textbf{92.45 $\pm$ 7.09} & \textbf{1.98 $\pm$ 0.00} & \textbf{88} \\
        \midrule
        40 & 430.57 $\pm$ 23.00 & 2.00 $\pm$ 0.05 & 34 & \textbf{125.28 $\pm$ 8.23} & \textbf{1.99 $\pm$ 0.02} & \textbf{82} \\
        \midrule
        50 & 588.84 $\pm$ 40.02 & 2.01 $\pm$ 0.08 & 12 & \textbf{144.53 $\pm$ 9.28} & \textbf{1.99 $\pm$ 0.03} & \textbf{76} \\     
        \midrule
        60 & - $\pm$ - & - $\pm$ - & 0 & \textbf{136.22 $\pm$ 10.32} & \textbf{1.99 $\pm$ 0.03} & \textbf{63} \\     
        \midrule
        70 & - $\pm$ - & - $\pm$ - & 0 & \textbf{142.54 $\pm$ 10.89} & \textbf{2.00 $\pm$ 0.03} & \textbf{54} \\     
        \midrule
        80 & - $\pm$ - & - $\pm$ - & 0 & \textbf{156.34 $\pm$ 11.44} & \textbf{2.00 $\pm$ 0.04} & \textbf{47} \\     
        \midrule
        90 & - $\pm$ - & - $\pm$ - & 0 & \textbf{168.56 $\pm$ 11.46} & \textbf{2.00 $\pm$ 0.04} & \textbf{42} \\     
        \midrule
        100 & - $\pm$ - & - $\pm$ - & 0 & \textbf{184.32 $\pm$ 12.05} & \textbf{2.00 $\pm$ 0.04} & \textbf{20} \\     
        \bottomrule
    \end{tabular}
\end{table*}

The principal factor contributing to these outcomes is the capacity of our algorithm to address the same problem while accounting for a reduced quantity of obstacles. \autoref{tab:activeres} presents the count of active obstacles within the solution framework during quadrotor testing. It is evident from the data that the average count of active obstacles across 100 randomly conducted tests significantly diminishes in comparison to the original count. Furthermore, in all instances, there exist scenarios where the necessity for active obstacles is completely obviated, resulting in a count of zero active obstacles. Notably, the number of active obstacles within the solution escalates commensurate with the increasing number of obstacles, aligning with expected trends.

 \begin{table}
     \centering
   \footnotesize 
   {%
   \renewcommand{\tabcolsep}{5.6pt} 
   \renewcommand{\arraystretch}{0.8} 
     \caption{The number of active obstacles in the solution in quadcopter scenarios across various obstacle quantities.}
     \label{tab:activeres}}
     \begin{tabular}{ cccc  }
      \toprule
      \multirow{2}{4em}{$n_{obs}$} & \multicolumn{3}{c}{Active Obstacle}   \\
     
      & mean & min & max \\
      \midrule
        5  & 0.14 & 0 & 2 \\
         10  & 0.35 & 0 & 2 \\
         15  & 0.54 & 0 & 4 \\
         20  & 0.97 & 0 & 5 \\
         25  & 1.08 & 0 & 7 \\
         30  & 1.13 & 0 & 5 \\
         40  & 2.01& 0 & 11 \\
         50  & 2.68 & 0 & 10 \\
         60  & 2.80 & 0 & 8 \\
         70  & 4.45 & 0 & 16 \\
         80  & 5.93 & 0 & 18 \\
         90  & 7.00 & 0 & 18 \\
         100  & 8.90 & 0 & 20 \\
        
     \bottomrule
     \end{tabular}

 \end{table}

\begin{figure*}
    \centering
    \includegraphics[width = .8\textwidth]{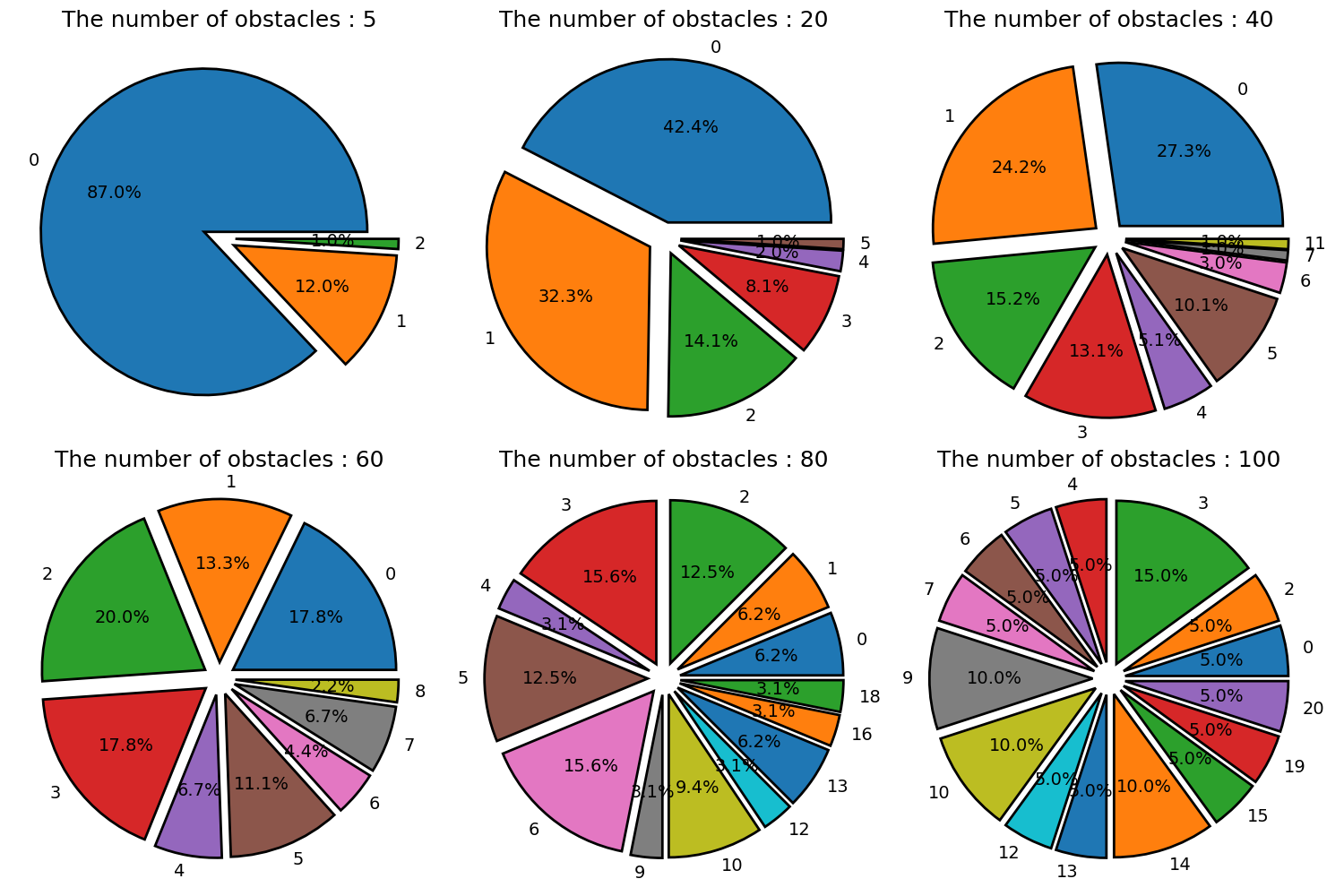}
    \caption{The distribution of the number of active obstacles in the solution for different numbers of obstacles in the quadrotor scenarios.}
    \label{fig:chart}
\end{figure*}

The graph presented in \autoref{fig:chart} illustrates the distribution of active obstacles within the solutions obtained from various quadrotor tests. In the case of scenarios involving five obstacles, it is noteworthy that approximately 87\% of the solutions are found to be devoid of any active obstacles, signifying a notable prevalence of obstacle-free solutions. However, this proportion experiences a diminishing trend as the number of obstacles increases. Notably, even in scenarios featuring a substantial 100 obstacles, a noteworthy 5\% of the solutions persist in not necessitating the presence of any active obstacles. Intriguingly, when the number of obstacles reaches 40, the occurrence of solutions without any active obstacles attains its highest rate.

\begin{figure*}
    \centering
    \includegraphics[width = \textwidth]{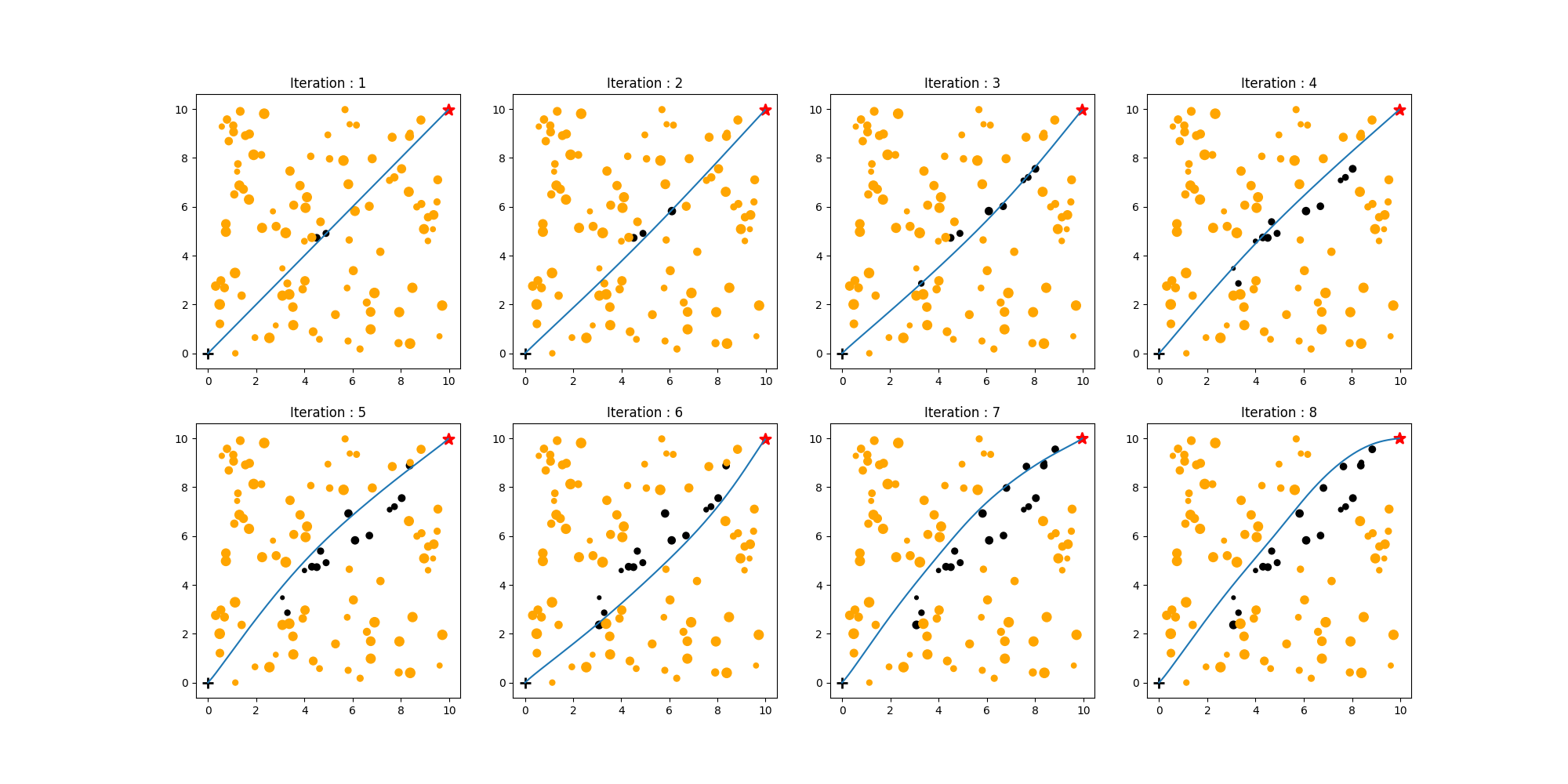}
    \caption{The findings pertain to a series of experimental scenarios involving 100 obstacles, within which the solution set encompasses 20 active obstacles. Notably, inactive obstacles are represented by the color orange, while active obstacles are denoted by the color black. The trajectory is illustrated as a blue line, with the initial point marked as '+' and the final point indicated by '*'. It is noteworthy that the longest consecutive solution observed in this study comprises eight iterative steps. Additionally, it is imperative to emphasize that each successive solution, with the exception of the final one, introduces a novel set of active obstacles to the ongoing optimization problem.}
    \label{fig:iter}
\end{figure*}

As delineated in \autoref{alg:method}, the proposed algorithm follows an iterative approach whose termination is contingent upon the specific characteristics of the problem under consideration. In our experimental evaluation, we observed that the highest number of iterations was necessitated when confronted with a scenario involving 100 obstacles. In this particular case, the algorithm converged after precisely 8 iterations, resulting in a trajectory that successfully circumvented 20 active obstacles, as illustrated in \autoref{fig:iter}.

As stipulated in the algorithmic framework, the initial iteration commences without any active obstacles, consequently yielding a trajectory that represents a straight line connecting the initial and final points. Subsequently, during the course of iterations, two obstacles were identified as compromising the feasibility of the trajectory. These obstacles were thus incorporated into the active obstacle set, a process that persisted as the algorithm iteratively adapted the trajectory to accommodate new obstacles. By the seventh iteration, the active obstacle set had expanded to include 20 distinct obstacles, prompting the algorithm to recalculate the trajectory. The result of this eighth iteration, which manifested a collision-free trajectory, was deemed the solution to the problem at hand.

\newpage

\newpage\phantom{blabla}

\section{Conclusions\label{sec:conclusion}}

In this paper, we proposed an iterative active-inactive obstacle approach for time-optimal obstacle avoidance. The proposed method involves iteratively identifying the active obstacles and recalculating the optimal path until a satisfactory solution is found. The effectiveness of the proposed method was demonstrated with two different dynamic models using a variety of obstacle numbers. The results showed that the proposed method was able to find the optimal path in a timely manner, while also being able to handle a large number of obstacles in the environment and the constraints on the motion of the object. The proposed method is a promising approach for time-optimal obstacle avoidance in complex environments. Future work could focus on improving the efficiency of the method, as well as extending it to handle more complex obstacle models.


\balance

\bibliographystyle{IEEEtran}

\end{document}